\documentclass[acmtog]{acmart}
\usepackage{booktabs}    
\usepackage{pdfpages}    
\usepackage{amsmath, bm} 
\usepackage{textcomp}
\usepackage{multirow}
\usepackage{multicol}
\usepackage{makecell}
\usepackage{relsize}
\usepackage{siunitx}
\usepackage[utf8]{inputenc}
\newcommand{\minus}{\scalebox{0.75}[1.0]{$-$}}

\acmSubmissionID{346}

\setcopyright{rightsretained}
\acmJournal{TOG}
\acmYear{2020}
\acmVolume{39}
\acmNumber{4}
\acmArticle{38}
\acmMonth{7}
\acmDOI{10.1145/3386569.3392433}

\setcitestyle{square}

\begin{document}

\title{\textbf{CARL}: \textbf{C}ontrollable \textbf{A}gent with \textbf{R}einforcement \textbf{L}earning for Quadruped Locomotion}

\author{Ying-Sheng Luo}
\email{luo.ying-sheng@inventec.com}
\authornote{Joint first authors.}
\affiliation{
    \institution{Inventec Corp.}
    \country{Taiwan}}

\author{Jonathan Hans Soeseno}
\email{soeseno.jonathan@inventec.com}
\authornotemark[1]
\affiliation{
    \institution{Inventec Corp.}
    \country{Taiwan}}

\author{Trista Pei-Chun Chen}
\email{chen.trista@inventec.com}
\affiliation{
    \institution{Inventec Corp.}
    \country{Taiwan}}
    
\author{Wei-Chao Chen}
\email{weichao.chen@skywatch24.com}
\affiliation{
    \institution{Skywatch Innovation Inc.}}
\affiliation{
    \institution{Inventec Corp.}
    \country{Taiwan}}
    
\begin{abstract}
Motion synthesis in a dynamic environment has been a long-standing problem for character animation.
Methods using motion capture data tend to scale poorly in complex environments because of their larger capturing and labeling requirement.
Physics-based controllers are effective in this regard, albeit less controllable.
In this paper, we present CARL, a quadruped agent that can be controlled with high-level directives and react naturally to dynamic environments.
Starting with an agent that can imitate individual animation clips, we use Generative Adversarial Networks to adapt high-level controls, such as speed and heading, to action distributions that correspond to the original animations. 
Further fine-tuning through the deep reinforcement learning enables the agent to recover from unseen external perturbations while producing smooth transitions.
It then becomes straightforward to create autonomous agents in dynamic environments by adding navigation modules over the entire process.
We evaluate our approach by measuring the agent's ability to follow user control and provide a visual analysis of the generated motion to show its effectiveness.
\end{abstract}



\begin{teaserfigure}
  \centering
  \includegraphics[width=7in]{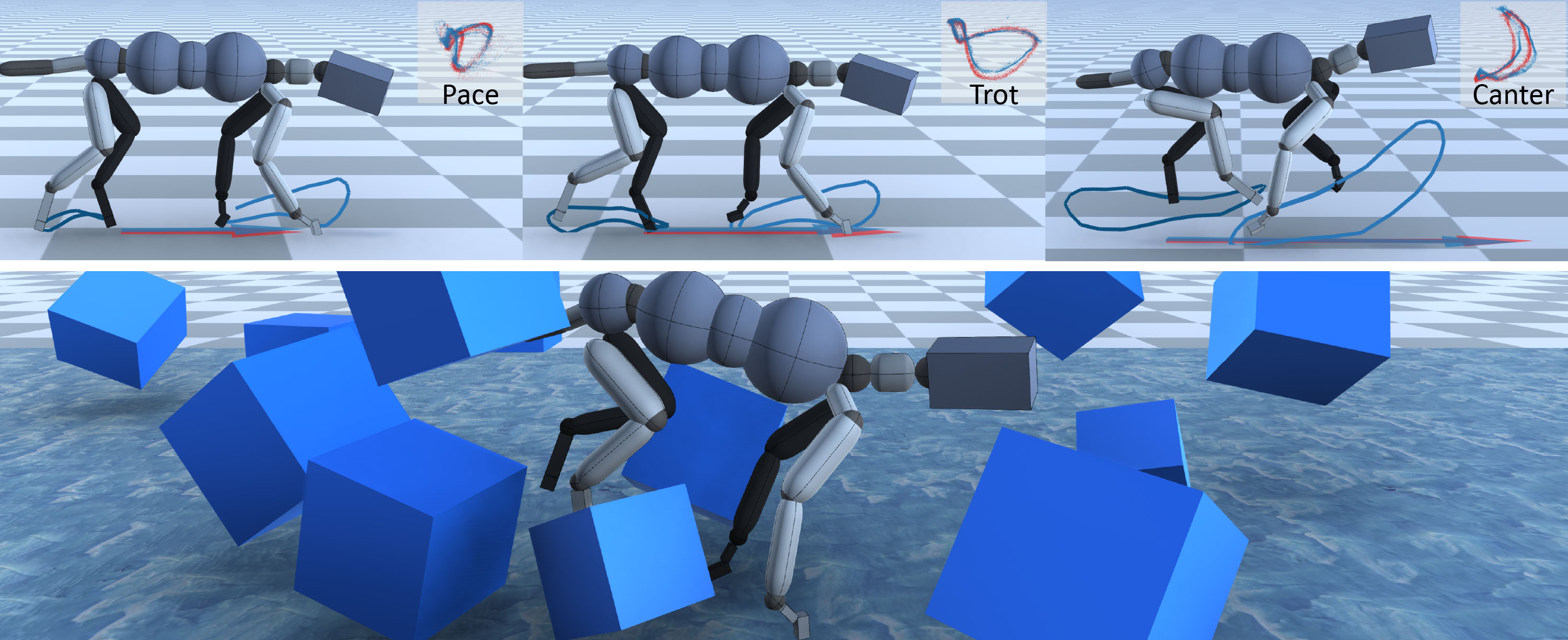}
  \caption{Our controllable agent produces natural movements (blue) alike depicted in the reference motion clip (red) and adapt to external perturbations.}
\end{teaserfigure}
\maketitle
\section{Introduction}

The quality of character animation in cartoons, video games, and digital special effects have improved drastically in the past decades with new tools and techniques developed by researchers in the field. Amongst various types of characters, quadrupeds are especially challenging to animate due to their wide variations of style, cadence, and gait pattern.  For real-time applications such as video games, the need to react dynamically to the environments further complicates the problem.

Traditionally, to synthesize new animations from motion capture data, one would create an interpolation structure such as a motion graph~\cite{Kovar2002motion,lee2002interactive}, where the nodes represent well-defined actions from motion capture data, and the edges define the transition between the actions. Aside from the long and tedious process of labeling the graph, it is often difficult to acquire sufficient motion capture data for quadrupeds to cover different gait patterns and styles. Furthermore, the motion graph would become impractically big and complex in dynamic environments to take into account the numerous interactions between the agent and its surroundings.  Despite the complexity, the motion graph would still not be useful for motion synthesis when unseen scenarios arise.

Research on the kinematic controller solves the labeling problem by reducing the need for crafting transitions between actions while allowing users to control the agent to produce the desired motions~\cite{zhang2018mann}. But since a kinematic controller is designed to imitate the motion dataset, the agent would fail to respond naturally when it encounters unseen interactions between the agent and its surroundings in dynamic environments. For example, in a scenario involving a quadruped agent walking on a slippery, undulating boat, it would clearly be highly impractical to collect, or hand-engineer, enough reference motions to train the kinematic controllers. One can certainly resort to physics-based controllers to model complex phenomenons effectively, as a physical simulation enables the agent to produce meaningful reactions to external perturbations without the need to collect or animate such a reaction. Although, physical constraints such as gravity, friction, and collision introduce numerous difficulties in designing a physics-based controller.

In this paper, we propose a \textbf{data-driven, physics-based, controllable} quadruped agent by adopting (1) the ability to interact physically with the dynamic environments, and (2) the natural movements learned from reference motion clips. 
It is a three-stage process that starts by learning the reference motions through imitation. The agent then learns to map high-level user controls such as speed and heading into joint actions with Generative Adversarial Networks (GANs). Through Deep Reinforcement Learning (DRL), the agent then gains the ability to adapt and recover from unseen scenarios, allowing us to synthesize meaningful reactions against external perturbations. One can then control the trained agent by attaching navigation modules that emulate higher-level directive controls, such as path-finding and ray-sensor. We summarize our contributions as follows:
\begin{itemize}
    \item A GAN supervision framework to effectively map between high-level user controls and the learned natural movements, 
    \item A physics-based controller for quadruped agents trained through DRL that can adapt to various external perturbations while producing meaningful reactions without the need for action labels, and
    \item A methodology to attach high-level navigation modules to the agent for tackling motion synthesis tasks involving agent-environment dynamic interactions.
\end{itemize}
\raggedbottom
\section{Related Works}
Real-time quadruped motion synthesis not only needs to consider the wide range of variations in gait patterns but also dynamic and unseen scenarios that happen in real-time. Works in the topic of real-time quadruped motion synthesis can be divided into two main categories: kinematic controllers and physics-based controllers. Recent advances in deep learning have also been incorporated into both categories of methods.

\subsection{Kinematic Controllers}
There has been an extensive amount of research works on character animation using kinematic-based methods. Classical approaches~\cite{kovar2004automated, safonova2007construction, Kovar2002motion, lee2002interactive} constructed motion graphs that included edges corresponding to each short motion clip. The resulting motions were then synthesized by graph searching. Alternatively,~\cite{grochow2004style, levine2012continuous, shin2006motion, chai2005performance} enabled smooth transitions through low-dimensional exploration. \citet{huang2013real} used an asynchronous time warping approach to handle the gait transitions of a quadruped agent. Despite their successes, motion graphs either require a large memory footprint, high order computation time, or extensive preprocessing, thus limiting the scalability to more complex movements or even larger datasets.

Recent approaches use deep learning that is known to scale better with a larger dataset. Recurrent Neural Network based approaches~\cite{fragkiadaki2015recurrent, zhou2018auto, lee2018interactive} predicted the future states given previous observations. \citet{holden2016deep} combined a standard feed-forward neural network with an auto-encoder to produce motions with a high degree of realism. In their later work,~\cite{holden2017phase} used a neural network to model the phase function of a biped agent, enabling the agent to traverse uneven terrains. However, their approach required extensive labeling of the agent’s phase. Exploiting mixtures of experts model to learn quadruped locomotion, Mode-Adaptive Neural Networks~\cite{zhang2018mann} suppressed the need for such a labeling process. Other recent works from~\cite{lee2018interactive, starke2019neural} could produce less foot sliding artifact while interacting with the environment but sacrifice the control responsiveness. 

In general, kinematic controllers scale with the quality of the motion clip dataset and often generate motions with higher-quality compared to physics-based controllers. However, their ability in producing dynamic movements where complex environments are presented is limited by the availability of such motions in the dataset. Capturing and hand-engineering enough motion data to accommodate a large number of possible agent behaviors are impractical.

\subsection{Physics Based Controllers}
Physics-based methods offer an effective way to model complex scenarios. Such types of controllers not only require to follow control commands but also need to maintain balance. Early works in locomotion controller design required human-insight ~\cite{yin2007simbicon, coros2009robust, coros2010generalized, coros2011locomotion}. \citet{lee2010data} designed a locomotion controller for biped agents by tracking modulated reference trajectories from motion capture data. \citet{ye2010optimal} designed an abstract dynamic model with hybrid optimization processes to achieve robust locomotion controllers. Alternatively, works in value iteration~\cite{coros2009robust} developed task-based controllers for biped agents with varying body proportions. \citet{peng2015dynamic} also used value iteration to produce a quadruped locomotion controller in 2D that could traverse terrain with gaps. However, their approach required manual selection of important features such as desired joint angles and leg forces. Although these approaches were robust, they did not scale with complex controls such as stylization or other high-level directive controls.

Trajectory-based approaches have also been explored in the past few years~\cite{liu2015improving, liu2010sampling, wampler2009optimal}. \citet{liu2010sampling} developed an open-loop controller for biped agents. These approaches did not scale well with complex and long motions. Such a problem was later solved by~\cite{liu2015improving} through the reconstruction of past experiences. Recent works on biped and quadruped agents~\cite{wampler2009optimal, wampler2014generalizing} explored the gait patterns of various legged animals. \citet{levine2012physically} produced plausible variations of different locomotion skills such as walking, running, and rolling. These approaches tended to struggle with long-term planning. Recent works~\cite{hamalainen2015online, tassa2012synthesis, tassa2014control} extended the offline approach to online optimization methods. Optimizing over short predictive horizons produced robust controllers~\cite{da2008simulation, tassa2012synthesis}. Physics-based controllers are considered an effective way to model complex phenomenons, because they produce novel movements from its interactions with the environment. However, incorporating high-level controls in physics-based controllers is not a trivial task and often causes the controllers to look unnatural or is limited in its movement diversity.
\raggedbottom

\subsection{Deep Reinforcement Learning (DRL)}
Physics-based methods with deep neural networks, esp., Deep Reinforcement Learning (DRL),~\cite{brockman2016openai, liu2017learning, peng2016terrain, bansal2017emergent, heess2017emergence, schulman2015trust, schulman2017ppo, sutton1998introduction, merel2018neural} trained controllers capable of performing diverse motions. Works from ~\cite{hwangbo2019learning, won2019learning} shown the success in controlling robots or agents with different morphology. Aerial locomotion was also explored through DRL algorithms~\cite{won2017train, won2018aerobatics}. Works in imitation learning~\cite{chentanez2018physics, peng2018deepmimic} produced high-quality motions through imitating well-defined short-clips. However, these methods' ability to scale with the size of motion capture data was limited.

Recent works from \cite{park2019learning} and \cite{bergamin2019drecon} combined kinematic controllers with DRL to produce a responsive controller for biped agents. Multiplicative compositional policies (MCP) from~\cite{peng2019mcp} adopted a new way to blend low-level primitives to follow the high-level user intent. Despite their successes with biped controllers, these techniques could not generalize to quadruped agents. As discussed in~\cite{zhang2018mann}, modeling gait transitions of quadruped agents were harder compared to biped agents. This was because of the gait pattern complexity and the limited number of available motion capture data.
\raggedbottom

\subsection{Adversarial Learning on DRL}
The main challenge of physics-based controllers is to produce natural movements. Defining a reward function that captures a movement's naturalness is difficult. Optimizing for high-level objectives such as maintaining a certain speed or following a certain turning angle does not guarantee the resulting motion being smooth or natural. Generative adversarial imitation learning (GAIL)~\cite{ho2016generative}, incorporated a neural network that provided supervision over the agent's movements, allowing the controller to learn the manifold of natural movements. Such an adversarial approach could produce robust controllers~\cite{fu2017learning}, yet the generated motions were still not comparable to the kinematic controllers. The framework of generative adversarial networks (GANs) is known to be a powerful estimator of a prior known distribution~\cite{choi2018stargan, zhu2017cyclegan, engel2019gansynth}. But it is also known to be unstable and sensitive to the objective function. Besides, a good objective function for natural gait transitions is difficult to define. In summary, DRL approaches struggle with defining an objective function that describes natural movements, directly adding GAN into the mix would disrupt the learning process, causing the controller to produce awkward movements.

\begin{figure*}[t]
  \centering
  \includegraphics[width=7.0in]{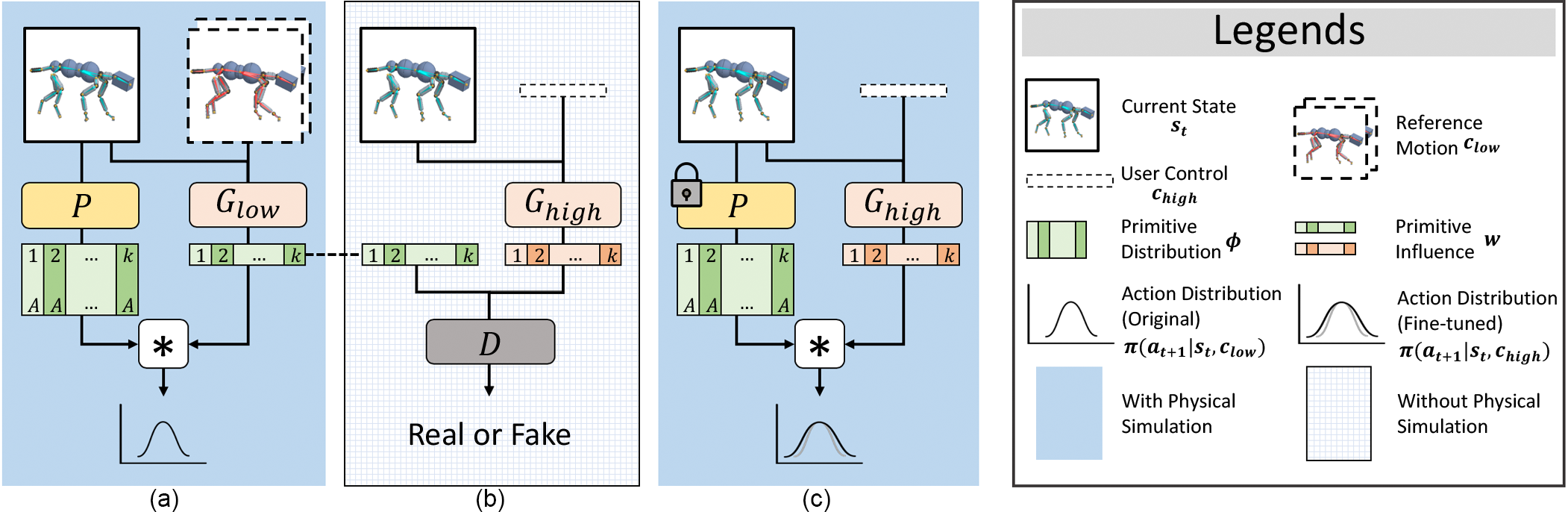}
  \vspace{-0.2cm}
  \caption{Our system is divided into three modules: (a) Low-Level: Imitation Learning, (b) High-Level: GAN Control Adapter, and (c) High-Level: DRL Fine-Tuning. During the fine-tuning process, we fixed the primitive network to prevent it from compensating the high-level gating network's error.}
  \label{fig:schematic_diagram}
  \vspace{-0.25cm}
\end{figure*} 

\section{Proposed Method}
\label{section:proposed_method}
Our goal is to design a physics-based controller that produces natural movements and reactions under external perturbations while following high-level user controls. We consider the generated motion as natural when it resembles the movement of the reference motion. For this purpose, we train the controller in three stages (Figure~\ref{fig:schematic_diagram}). In the first stage, we aim to transfer the natural movements from the reference motion clips to a physics-based controller through imitation learning (Figure~\ref{fig:schematic_diagram}(a)). This is achieved by learning the action distribution through a policy network for the physics-based controller to follow. This policy network contains a primitive and gating network that decompose the action distribution into lower-level primitive distributions that can be multiplicatively composed using learned weightings. As a result, the policy network produces an action distribution that enables the physics-based controller to produce natural movements, successfully bridging animation and physics.

To enable the controller to adopt high-level control objectives such as target speed and heading, it would not be feasible to directly optimize for forward progression, as this would end up producing a different action distribution which may result in awkward and unnatural movements. Therefore, in the second training stage (see Figure \ref{fig:schematic_diagram}(b)), we adopt a GAN Control Adapter to enable the high-level gating network to approximate the natural action distribution learned previously. However, the lack of external perturbations during the second training stage hinders the controller's ability to adapt to unseen scenarios. Therefore, we add a GAN regularized DRL fine-tuning stage to empower the controller to recover from such scenarios. We further discuss the details of each training stage in Section~\ref{sec:low_level_imitation}, Section~\ref{sec:high_level_offlinegan}, and Section~\ref{sec:high_level_drl}.

\subsection{Low-Level: Imitation Learning}
\label{sec:low_level_imitation}

Our goal in this stage is to learn the action distribution for a physics-based controller. To do so, we need the action distribution to describe the articulation of the agent. This can be accomplished through imitation learning where we treat the reference motion as a form of low-level control that specifies the agent's movement at the joint level. For this purpose, we adopt the policy network of~\cite{peng2019mcp} which consists of two modules: gating and primitive networks (see Figure.~\ref{fig:schematic_diagram}(a)). The low-level gating network $G_{\text{low}}$ takes the agent's current state $s_t$ and the reference motion $c_{\text{low}}$ as input and produces primitive influence $\textbf{w}$,

\begin{equation}
    \textbf{w} = G_{\text{low}}(s_t, c_{\text{low}}), 
\end{equation}

where $\textbf{w} \in \mathbb{R}^{k}$, with k being the total number of primitive motions and $c_{\text{low}} = (\hat{s}_{t+1}, \hat{s}_{t+2})$ is the joint-level control defined as the target states for the next two time steps of the reference motion. Both $s_t$ and $\hat{s}_t$ contain the information of each joint's position, velocity, rotation, and angular velocity. All the information are represented as 3-dimensional vectors except for the rotations, which are represented as 4-dimensional quaternions. The redundant storage of both position and rotation in the state representation is a widely adopted technique for learning the locomotion skills for torque-articulated characters~\cite{todorov2012mujoco, peng2018deepmimic, park2019learning}.

The primitive network $P$, on the other hand, takes in the agent's current state $s_t$ and decomposes the action distribution into primitive distributions $\phi_1 ... \phi_k$, where each primitive $\phi_i$ represents the lower-level action distribution that specializes in modeling a specific locomotion skill within the action space. Each $\phi_i$ is modeled as a Gaussian distribution with state-dependent action mean $\mu_i(s_t)$ and diagonal covariance matrix $\Sigma_i$, 

\begin{equation}
    \phi_{i} = \mathcal{N}(\mu_i(s_t), \Sigma_i), {i = 1, 2 ... k}.
\end{equation}

Unlike~\cite{peng2019mcp} which learns the state-dependent covariances, we use a fixed diagonal covariance matrix $\Sigma$. We found that learning $\Sigma$ leads to premature convergence, similar to the findings of~\cite{merel2018hierarchical}. As discussed in~\cite{peng2019mcp}, multiplicatively composing the Gaussian primitive distributions $\phi$ given its influence $\textbf{w}$ produces a composite distribution that is also a Gaussian distribution

\begin{equation}
    \pi(a_{t+1}|s_{t},c_{t}) = \frac{1}{Z(s_t,c_t)} \mathlarger{\Pi_{i=1}^{k}} \phi_i^{w_{i}}, 
\end{equation}

where $Z(s_t,c_t)$ denotes a normalization function, and $c_t$ denotes the current control objective $c_t=c_{\text{low}}$. The agent's next action $a_{t+1}$ is then sampled from this action distribution. Following ~\cite{peng2018deepmimic}, we use their pose, velocity, and center-of-mass (COM) rewards:

\begin{equation}
    R_{\text{p}} = \exp[-2\big(\sum_j||\hat{q}_j\ominus q_j||^2 \big)],
\end{equation}
\begin{equation}
    R_{\text{v}} = \exp[-0.1\big(\sum_j ||\hat{\dot{q}}_j-\dot{q}_j||^2 \big)], 
\end{equation}
\begin{equation}
    R_{\text{com}} = \exp[-10\big(||\hat{p}_{\text c}\ - p_{\text c}||^2 \big)].
\end{equation}
The pose reward $R_{\text{p}}$ encourages the controller to match the reference motion's pose by computing the quaternion difference  $\ominus$ between the agent's $j\text{-}th$ joint orientations $q_j$ and the target $\hat{q}_j$. The velocity reward $R_{\text{v}}$ computes the difference of joint velocities, where $\dot{q}_j$ and $\hat{\dot{q}}_j$ represent the angular velocity for the $j\text{-}th$ joint of the agent and the target respectively. The center-of-mass reward $R_{\text{com}}$ discourages the agent's center-of-mass $p_{\text c}$ to deviate from the target's $\hat{p}_{\text c}$.

We replace the end-effector reward with contact point reward $R_{\text c}$,
\begin{equation}
    R_{\text c} = \exp[-\frac{\lambda_{\text c}}{4}\big(\sum_e\hat{p}_e\oplus p_e \big)], \lambda_{\text c} = 5, 
\end{equation}

where $\oplus$ denotes the logical XOR operation, $p_e$ denotes the Boolean contact state of the agent's end-effector, and $e \in\{$\textit{left-front}, \textit{right-front}, \textit{left-rear}, \textit{right-rear}$\}$. This reward function $R_{\text c}$ is designed to discourage the agent when the gait pattern deviates from the reference motion, and to help resolve the foot-sliding artifacts. For instance, the Boolean contact state for only the \textit{left-front} end-effector touches the ground can be denoted by $p = [1, 0 , 0, 0]$.  Inspired by~\cite{peng2018deepmimic}, we use the exponential value where $\lambda_{\text c}$ denotes a hyper-parameter that controls the slope of the exponential function, and found that $\lambda_{\text c} = 5$ yields the best outcome.  

\begin{equation}
\begin{aligned}
    R = 0.65R_{\text p} + 0.1R_{\text v} + 0.1R_{\text{com}} + 0.15R_{\text c},
\end{aligned}
\end{equation}

is the final form of the reward function. To reduce the training complexity, we separate the reference motion of each control objective (see Section~\ref{sec:details_online_drl}). As a result, our physics-based controller can imitate the given reference motion by learning the corresponding action distribution. The controller produces natural movements, performing different gait patterns depicted by the reference motion.

\subsection{High-Level: GAN Control Adapter}
\label{sec:high_level_offlinegan}

Our goal for this stage is to enable high-level user control over speed and heading. This allows the user to directly control the agent without going through the tedious and laborious process of specifying reference motions. However, directly optimizing for forward progression or high-level user controls (such as target speed $R_{\text{spd}}$ or heading $R_{\text{head}}$ defined later in Eq.~\ref{eq:high_speed} and Eq.~\ref{eq:high_heading}, respectively) causes the policy network to ignore the previously learned action distribution, and this eventually leads to awkward and unnatural movements. For the remainder of our discussion, we represent the high-level user control with two values $c_{\text{high}} = (\sigma, \Delta\theta)$, where $\sigma$ denotes the agent's target speed (m/s), and $\Delta\theta$ denotes the angular difference between the agent's current heading and target heading. For instance, the control for the agent to travel at 1 m/s while rotating 90 degrees counter-clockwise is $c_{\text{high}} = (1, 0.5\pi)$. During training, we consider $c_{\text{low}}$ and $c_{\text{high}}$ as paired labels where $c_{\text{high}}$ is derived from the physical states of the reference motion $c_{\text{low}}$.

Since replacing the reference motion with high-level user controls only affects the gating network, learning the mapping between high-level user control and the primitive influence translates to learning the agent’s action distribution. Standard distance functions such as $l1$ or $l2$ only preserve the low-order statistics of a distribution and do not guarantee the samples are drawn from the correct distribution. Therefore, we use the GAN framework which, instead of estimating low-order statistics, approximates the manifold of the distribution through adversarial learning. The GAN framework consists of two modules: generator and discriminator. Given real samples of primitive influence $\textbf{w}_{\text{real}}$ drawn from real data distribution $\textbf{w}_{\text{real}} \sim G_{\text{low}}(s_t,c_{\text{low}})$, our high-level gating network $G_{\text{high}}$ serves as the generator and produces fake primitive influence $\textbf{w}_{\text{fake}}$ which is drawn from $\textbf{w}_{\text{fake}} \sim G_{\text{high}}(s_t,c_{\text{high}})$. Our discriminator $D$ aims to distinguish fake samples $\textbf{w}_{\text{fake}}$ from the real samples $\textbf{w}_{\text{real}}$ by maximizing $L_{\text{adv}}$ defined in Eq~\ref{eq:adv}. 

\begin{equation}
    \begin{aligned}
    \min_{G_{\text{high}}} \max_{D} L_{\text{adv}} = & \mathbb{E}_{s_t,c_{\text{low}}}[\log D(G_{\text{low}}(s_t,c_{\text{low}}))] + \\
               & \mathbb{E}_{s_t,c_{\text{high}}}[\log (1 - D(G_{\text{high}}(s_t,c_{\text{high}})))]
    \end{aligned}
    \label{eq:adv}
\end{equation}

\begin{equation}
    L_{\text{rec}} = || \textbf{w}_{\text{fake}} - \textbf{w}_{\text{real}} ||_1
\end{equation}
\begin{equation}
    L_G = \lambda_{\text{adv}} * L_{\text{adv}} + \lambda_{\text{rec}} * L_{\text{rec}}
    \label{eq:lg}
\end{equation}

We train $G_{\text{high}}$ by minimizing the objective function $L_G$ defined in Eq.~\ref{eq:lg}. Reconstruction loss $L_{\text{rec}}$ minimizes the absolute distance between samples drawn from real and fake data distributions. Through the adversarial loss $L_{\text{adv}}$, the discriminator $D$ provides supervision to the generated samples by classifying it as real or fake. This guides $G_{\text{high}}$ to learn the real data distribution, i.e., the manifold. Inspired by~\cite{isola2017pix2pix}, we weigh the importance of each loss by $\lambda_{\text{rec}}=100$, and $\lambda_{\text{adv}}=1$. Jointly minimizing the two losses $L_{\text{rec}}$ and $L_{\text{adv}}$ allows the generator to produce fake samples that are close in terms of distance and also come from the real data distribution. The end result is a control adapter that translates high-level user control over target speed and heading into natural movements.

\begin{figure*}[ht!]
  \centering
  \includegraphics[width=7.0in]{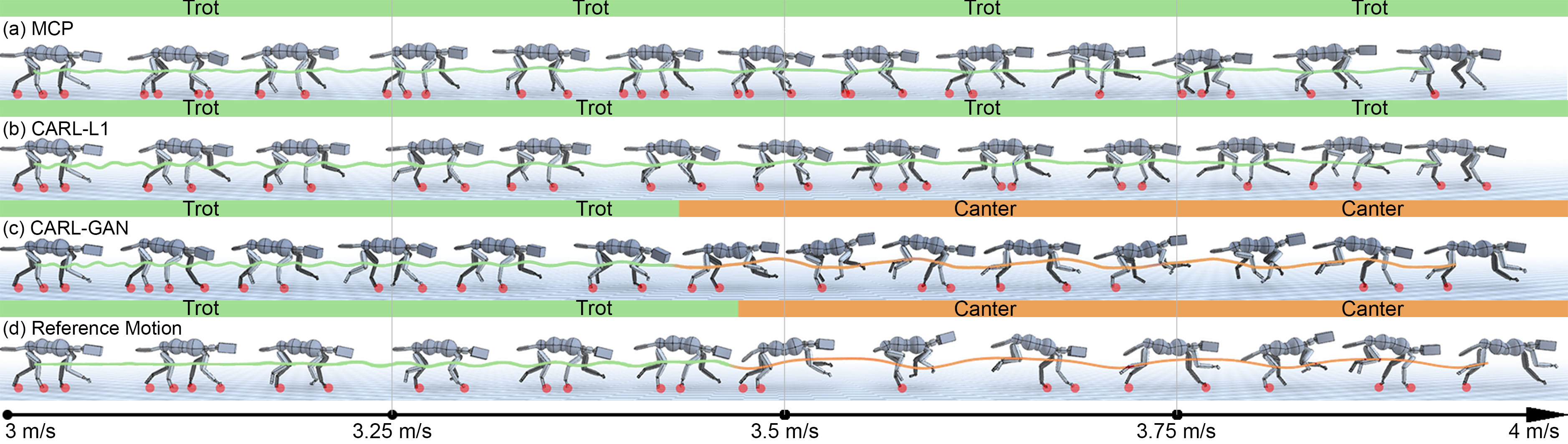}
  \vspace{-0.2cm}
  \caption{The agent follows the user speed control from 3-4 m/s in a straight direction. (a) MCP (No Control Adapter) and (b) CARL-L1 (L1 Control Adapter) continue to perform trot at unnaturally high cadence. Whereas the (c) CARL-GAN (Ours) and (d) Reference Motion both automatically switch gait pattern from trot to canter as the target speed becomes higher.}
  \label{fig:velocity_qual}
  \vspace{-0.35cm}
\end{figure*} 

\subsection{High-Level: DRL Fine-Tuning}
\label{sec:high_level_drl}

During the GAN training process, we only expose a small subset of possible scenarios to the controller. This means rarely sampled controls and external perturbations can cause the agent to fall, as the controller may be unable to recover from these unseen scenarios.  We solve this problem by further fine-tuning the agent with physical simulations, where we expose the agent to a significantly larger number of scenarios and ask it to recover through trial and error. For example, when being bombarded with objects or when the speed suddenly changes, the controller should learn to shift the agent's body weight to prevent it from falling.

We have observed the tendency for the policy network to change the primitive network to compensate for the gating network's error. As discussed previously in Section~\ref{sec:high_level_offlinegan}, since high-level user control only affects the primitive influence, we freeze the parameters of the primitive network and only train the gating network to preserve the action distribution. We train the high-level gating network with the corresponding reward function $R_{\text{spd}}$ or $R_{\text{head}}$ depending on the control objective.

\begin{equation}
    {R_{\text{spd}} = \text{exp}[-\lambda_{\text{spd}}(\sigma-||\textbf{v}||)^2]; \lambda_{\text{spd}} = 0.8}
    \label{eq:high_speed}
\end{equation}

\begin{equation}
    R_{\text{head}} = \bigg(\frac{\hat{\textbf{u}}\cdot \textbf{v}}{||\hat{\textbf{u}}||*||\textbf{v}||} + 1\bigg) * 0.5
    \label{eq:high_heading}
\end{equation}

The speed reward $R_{\text{spd}}$ computes the $l2$-distance between the target speed denoted by $\sigma$ and the agent's current movement speed $||\textbf{v}||$. We found that using $\lambda_{\text{spd}} = 0.8$ produces the best result in our case. As for the heading reward $R_{\text{head}}$, we compute the cosine similarity between the target heading $\hat{\textbf{u}} = (\cos(\hat{\theta}), \minus \sin(\hat{\theta}))$ and agent's heading $\textbf{v}$ projected onto the plane of motion, with $\hat{\theta}$ representing the target heading in radians. We normalize the value of cosine similarity between 0 and 1.

Lastly, to ensure the controller does not deviate too far from the learned action distribution, we impose a regularization term $L_{\text{reg}}$, 

\begin{equation}
    L_{\text{reg}} = \sum_{l=1}^{L}||\hat{\alpha}_l - \alpha_l||_1, 
    \label{eq:high_level_reg}
\end{equation}

where $\hat{\alpha}_l$ denotes the parameters of $l{\text -}th$ fully-connected layer of the GAN trained gating network (also used as the initialization point), $\alpha_l$ denotes the parameters of $l{\text -}th$ fully-connected layer of the currently trained gating network, and $L$ denotes the total number of layers in each gating network. We apply the regularization in the parameters space, because applying it to the layer's output would penalize the real unseen scenarios. After fine-tuning with DRL, our controller gains the ability to recover from unseen scenarios, enabling smooth transitions following user high-level controls while being adaptive to external perturbations.

\section{Implementation}
\subsection{Data Collection}
\label{sec:data_collection}

There are multiple ways to obtain the reference motion dataset, such as kinematic controllers, motion graphs, or even capturing raw motion data. For convenience, we use a readily available kinematic controller\footnote{https://github.com/sebastianstarke/AI4Animation} provided by \cite{zhang2018mann}. Since our goal is to control both speed and heading, collecting reference motion data containing all the possible movement speeds, turning rates, and turning angles would make the reference motion undesirably complex. As discussed by~\cite{peng2018deepmimic}, the complexity of the reference motion highly affects the success of the imitation learning process. We record two datasets, one for each speed and heading controls. For each control objective, we record one minute worth of reference motion clip. For speed control, the agent travels towards a fixed direction with various move speeds ranging from 0 to 4 m/s. The agent automatically changes its gait pattern depending on the speed. The agent performs pace at (0, 2) m/s, trot at [2, 3.5) m/s, and canter at [3.5, 4] m/s. As for heading control, the agent performs turning left and right in the range of 0 to 180 degrees. Here, the agent only performs pace gait-pattern as it moves at a fixed speed of 1 m/s.

\subsection{Online Optimization with DRL}
\label{sec:details_online_drl}
We train the policy network in a single PC equipped with AMD R9 3900X (12 Cores/24 Threads, clock speed at 3.8GHz). We adopt a physical simulation C++ Bullet physics library~\cite{coumans2013bullet} for physical simulation, which updates at 1200 frames per second. Our physical simulation modules query the policy network to obtain the action distribution at 30 frames per second. Our policy network learns with proximal policy optimization (PPO)~\cite{schulman2017ppo} and generalized advantage estimation (GAE)~\cite{schulman2015high}. We follow ~\cite{peng2019mcp}'s policy network architecture, number of primitives ($k=8$), and action space representation. The action at time $t$, $a_t \in A$, is represented as PD-targets placed at each joint. However, instead of learning the action-noise variance, we use a fixed diagonal covariance matrix $\Sigma=0.05I$, where $I$ is the identity matrix. The value function uses multi-step returns with TD($\lambda$)~\cite{sutton1998introduction}. The learning process is episodic, wherein each episode we incorporate early termination and reference state initialization proposed by~\cite{peng2018deepmimic}. 

In low-level imitation learning, we separate the learning process of speed and heading controls, each imitating the corresponding recorded reference motion clip as mentioned in Section~\ref{sec:data_collection}. We then have two separate policy networks: one for speed control and the other for heading control. Learning rates of the policy networks and value function networks are 2.5e-6 and 1e-2 respectively. Hyper-parameters for GAE($\lambda$), TD($\lambda$), and the discount factor are set to 0.95. The learning process takes approximately 250 hours for each policy network.

\begin{figure}[b!]
  \centering
  \vspace{-0.25cm}
  \includegraphics[width=.9\linewidth]{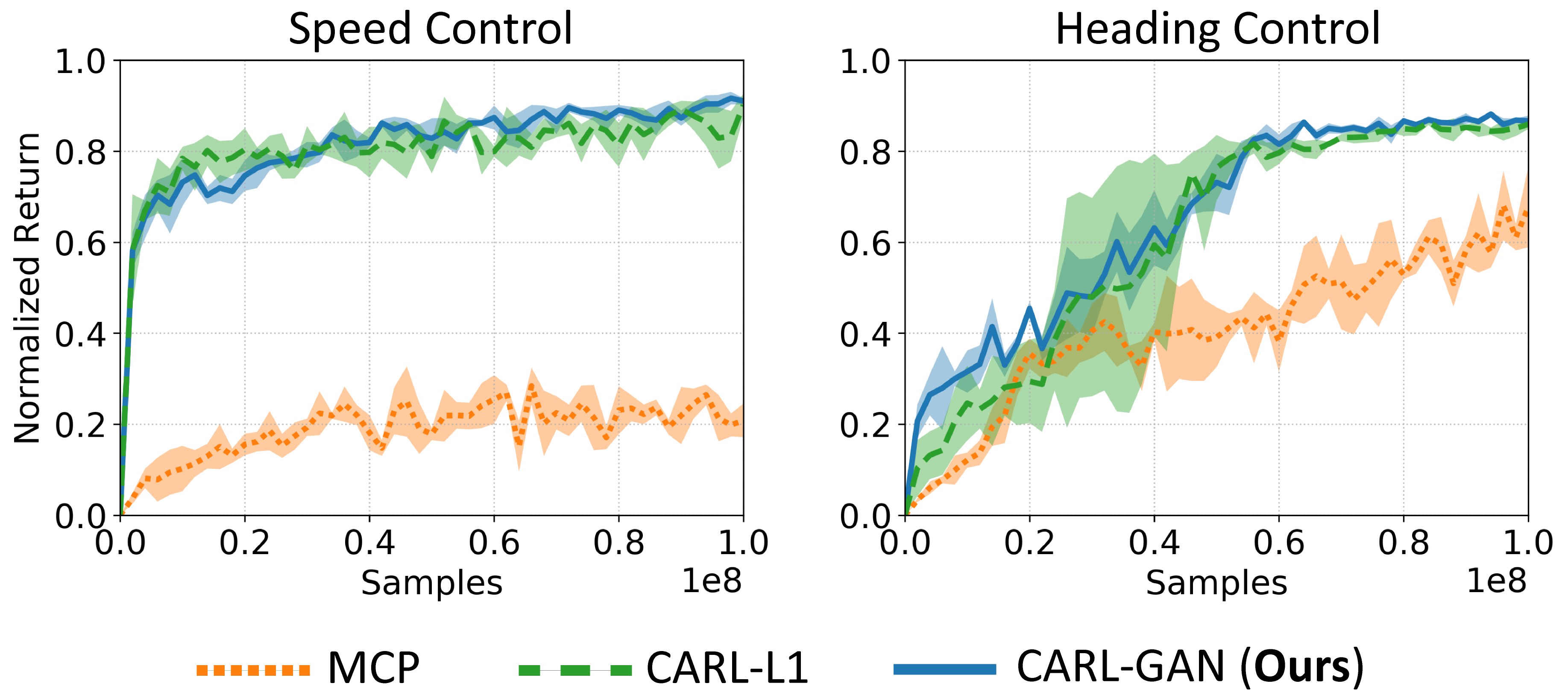}
  \vspace{-0.3cm}
  \caption{The learning curves of (left) speed and (right) heading control. MCP (No Control Adapter), CARL-L1 (L1 Control Adapter), and CARL-GAN (Ours).}
  \vspace{-0.1cm}
  \label{fig:learning_curve}
  \vspace{-0.25cm}
\end{figure}

In the high-level DRL fine-tuning step (Section~\ref{sec:high_level_drl}), the two policy networks optimize for different reward functions. The policy network for the move speed task uses $R_{\text{spd}}$ (defined in Eq.~\ref{eq:high_speed}) because the agent only moves towards a single direction. The policy network for the heading task only uses $R_{\text{head}}$ (defined in Eq.~\ref{eq:high_heading}) because the agent moves at the same speed. During training, we randomly update the control objective with an offset uniformly sampled from $[-0.25, 0.25]$ m/s and $[-0.15, 0.15]$ radian for speed and heading respectively. The updates happen at 4 frames per second. To encourage the controller to learn the rarely sampled gait transitions, we further introduce a 10\% probability that the control objective being completely altered (e.g. from cantering at 4 m/s to standing at 0 m/s). The learning rates for the policy networks and value function networks are 5e-5 and 1e-2 for both speed and heading respectively, with 0.99 for the discount factor. 

\begin{table}[t!]
    \caption{The MSE of speed at different gait patterns: pace, trot, and canter.}
    \vspace{-0.25cm}
    \centering
    \begin{tabular}{ |c|r@{\;{$\pm$}\;}r|r@{\;{$\pm$}\;}r|r@{\;{$\pm$}\;}r|} 
        \Xhline{2\arrayrulewidth}  
        \textbf{Gait} & \multicolumn{2}{c|}{\textbf{MCP}} & \multicolumn{2}{c|}{\textbf{CARL-L1}} & \multicolumn{2}{c|}{\textbf{CARL-GAN}} \\
        \Xhline{2\arrayrulewidth}
        Pace    & 1.7e-3  & 0.2e-3 & 17.7e-3 & 2.5e-3  & \textbf{1.6e-3} & \textbf{0.1e-3}\\
        \hline
        Trot    & 5.7e-3  & 1.9e-3 & 3.2e-3  & 0.8e-3  & \textbf{0.5e-3} & \textbf{0.3e-3}\\
        \hline
        Canter  & 16.4e-3 & 1.2e-3 & 28.1e-3 & 10.6e-3 & \textbf{4.1e-3} & \textbf{1.1e-3}\\ 
        \hline
        Avg.    & 7.9e-3  & 1.1e-3 & 16.3e-3 & 4.6e-3  & \textbf{2.1e-3} & \textbf{0.5e-3}\\         
        \Xhline{2\arrayrulewidth}
    \end{tabular}
    \label{tab:velocity_quant_mse}
    \vspace{-0.35cm}
\end{table}

The GAN regularization (defined in Eq.~\ref{eq:high_level_reg}) is added to the DRL's PPO clip surrogate loss as an additional term multiplied by 0.001. Our policy networks converge after approximately 25 hours for training the speed control and 40 hours for training the heading control. Alternatively, directly learning the high-level user control without GAN Control Adapter, the policy network converges after approximately 100 hours for training each of the control objectives. The learning curves are shown in Figure~\ref{fig:learning_curve}.

\subsection{Offline optimization with GAN}
The training process involves collecting one million data points consisting of the agent's state, reference motions, and high-level user controls (derived from two consecutive reference motions). During training, the real samples are drawn from the real data distribution by passing the agent' state and high-level user control to the low-level gating network.

We implement the discriminator as a series of 3-layered fully-connected neural networks with the LeakyReLU activation function in between the intermediate layers excluding the last classification layer. It optimizes for the original GAN objective where the discriminator outputs the probability of the input samples being real, as formulated by~\cite{goodfellow2014originalgan}. Inspired by~\cite{zhu2017cyclegan, isola2017pix2pix} we added the reconstruction loss $L_{\text{rec}}$ (defined in Eq.~\ref{eq:high_level_reg}). The training process converges in approximately 1 hour (50 epochs) using Adam optimizer with a learning rate of 2e-5.
\section{Experiment}
\label{sec:ablation_study}

In this section, we evaluate the effectiveness of using the GAN Control Adapter to enable user’s speed and heading controls over the agent. For this purpose, we consider two additional algorithms in addition to our proposed algorithm (\textbf{CARL-GAN}). Removing the GAN Control Adapter component of our method makes it equivalent to \textbf{MCP} \cite{peng2019mcp} that directly applies DRL for high-level user control. \textbf{CARL-L1} uses a standard distance function $L1$ instead of GAN as Control Adapter (i.e., L1 Control Adapter) to minimize the absolute distance between the output of the high-level gating network and the previously learned primitive influence.

Our evaluations focus on measuring the controller's ability to follow the user control in addition to the quality of the motion. We use the motions generated from the kinematic controller~\cite{zhang2018mann} as the ground truth. To measure the motion quality, we use an approach from robotics~\cite{moro2012kinematic, dholakiya2019design}, where the range of joint movements defines motion similarity. The visual results are shown in Figure \ref{fig:velocity_qual}.

Next, we discuss the quantitative results. A controller’s ability to follow user control is shown in Section \ref{sec:ablation_accuracy}; quality of generated motions is shown in Section \ref{sec:ablation_quality}; visualization and exploration of the action distribution are shown in Section \ref{sec:latent_space_exploration}; and discussion about our advantages compared to the Kinematic Controller + DRL approach in Section \ref{sec:kcdrl}.

\begin{table}[t!]
    \caption{Deviations from the reference motions, measured in degree ($^{\circ}$) for angular deviation, and meter (m) for positional deviation.}
    \vspace{-0.25cm}
    \centering 
    \def\arraystretch{1.1}
    \begin{tabular}{ |c|c|c|r@{\;$\pm$\;}r|r@{\;$\pm$\;}r|r@{\;$\pm$\;}r| } 
    \Xhline{2\arrayrulewidth}
    \textbf{Type} & \multicolumn{2}{c|}{\textbf{Control}} & \multicolumn{2}{c|}{\textbf{MCP}} & \multicolumn{2}{c|}{\textbf{CARL-L1}} & \multicolumn{2}{c|}{\textbf{CARL-GAN}}\\
    \Xhline{2\arrayrulewidth}
    \multirow{4}{*}{\rotatebox[origin=c]{90}{Angular ($^{\circ}$)}}  & \multirow{2}{*}{180$^{\circ}$} 
    & L  & 53.68 & 6.41 & 17.93 & 1.50  & \textbf{14.87} & \textbf{0.75}\\ &    
    & R  & 64.30 & 8.32 & 26.78 & 12.63 & \textbf{15.36} & \textbf{0.65}\\ \cline{2-9}
    & \multirow{2}{*}{90$^{\circ}$}   
    & L  & 46.27 & 2.60 & 18.05 & 1.23  & \textbf{15.42} & \textbf{0.60}\\ &    
    & R  & 44.54 & 3.03 & 22.34 & 3.03  & \textbf{13.04} & \textbf{1.28}\\ \Xhline{2\arrayrulewidth}
    \multirow{4}{*}{\rotatebox[origin=c]{90}{Position (m)}} & \multirow{2}{*}{180$^{\circ}$} 
    & L  & 0.95  & 0.06 &  0.31 & 0.03  & \textbf{ 0.25} & \textbf{0.05}\\ &
    & R  & 1.18  & 0.08 &  0.54 & 0.21  & \textbf{ 0.19} & \textbf{0.07}\\ \cline{2-9}
    & \multirow{2}{*}{90$^{\circ}$}  
    & L  & 1.26  & 0.07 &  0.31 & 0.04  & \textbf{ 0.21} & \textbf{0.01}\\ &
    & R  & 1.31  & 0.09 &  0.38 & 0.07  & \textbf{ 0.19} & \textbf{0.01}\\
    \Xhline{2\arrayrulewidth}
    \end{tabular}
    \label{tab:direction_quant}
    \vspace{-0.35cm}
\end{table}

\subsection{Control Accuracy}
\label{sec:ablation_accuracy}

To evaluate the controller’s ability in following commands, we produce 10 recordings, each with 5 seconds of duration, for each control objective. For speed control, the agent travels in a fixed direction with different gait patterns, e.g., pace, trot, and canter. Similarly for heading control, the agent travels at a constant speed (1 m/s) while performing 90- and 180-degree turns, both clockwise and counter-clockwise.  

\paragraph{Speed Control.}

Table~\ref{tab:velocity_quant_mse} shows our algorithm’s ability to follow the target speed at different gait patterns. The results show a significantly lower mean squared error (MSE) of our algorithm CARL-GAN compared to both baselines: MCP and CARL-L1. As expected, learning the target speed directly (MCP) allows the algorithm to converge better compared to L1 Control Adapter (CARL-L1). Note that the MSE of the canter gait pattern is greater than other gait patterns across all algorithms, due to its larger range of motion.

\paragraph{Heading Control.}

To investigate the controller’s responsiveness to the user’s heading control, we ask the controller to follow pre-defined trajectories of turning left-$90^\circ$, right-$90^\circ$, left-$180^\circ$, right-$180^\circ$, and calculate their respective angular and positional deviations from the reference motions (Table~\ref{tab:direction_quant}).

MCP's controller ignores the previously learned gait patterns and instead overfits the target heading, resulting in relatively high deviations. CARL-L1 performs markedly better than MCP because the control adapter preserves the learned gait patterns. Results from CARL-L1 is a bit biased with better MSE in left turns than right turns, due to slight left-turn and right-turn training data imbalance. Finally, our proposed algorithm CARL-GAN performs the best in all turning angles. It handles turns well even for drastic angles while following the target speed.   

\subsection{Motion Quality}
\label{sec:ablation_quality}
\begin{figure}[t!]
  \centering
  \includegraphics[width=0.99\linewidth, clip]{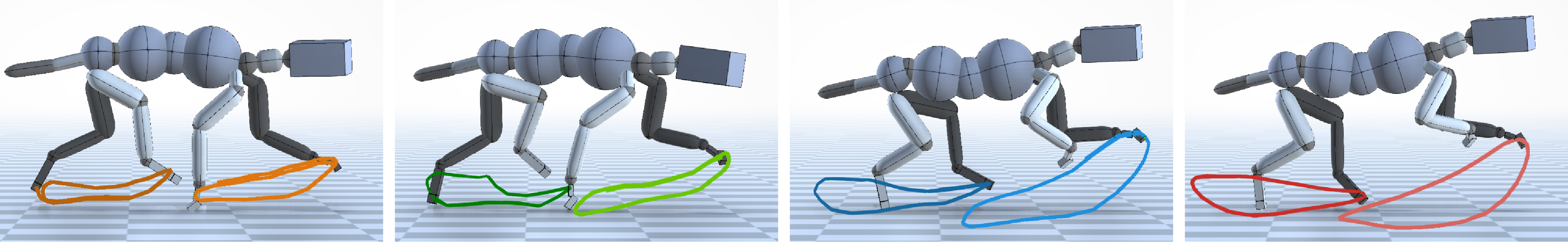}
  \vspace{-0.2cm}
  \caption{An end-effector moves in a cycle. Left-to-right: MCP (No Control Adapter), CARL-L1 (L1 Control Adapter), CARL-GAN (Ours), and Reference Motion.}
  \label{fig:end_effector_region}
  \vspace{-0.35cm}
\end{figure}

To measure the quality of the generated motions, we ask the controller to perform one gait pattern at a time and record its range of movements in the form of end-effector regions  (Figure~\ref{fig:end_effector_region}).
Along the line of robotics research~\cite{moro2012kinematic, dholakiya2019design}, we measure the similarity of movements by calculating the Intersection over Union (IoU) of the end-effectors between the generated motion and the reference motion (Table~\ref{tab:velocity_quant_iou}).
The results show that our controller achieves the highest IoU score consistently across gait patterns.

These quantitative results translate fairly well to the visual motion quality in Figure~\ref{fig:velocity_qual}.
For a reference motion that transitions from trot to canter (Figure~\ref{fig:velocity_qual}(d)), the controllers in both MCP and CARL-L1 continue to perform trot at unnaturally higher cadences instead of switching to canter gait-pattern.
In contrast, our controller can map the high-level user speed control to the previously learned action distribution, performing trot-to-canter transition correctly with natural movements. The high IoU score when performing canter in Table \ref{tab:velocity_quant_iou} further highlights this property. 

\begin{figure}[t!]
  \centering
  \includegraphics[width=0.99\linewidth]{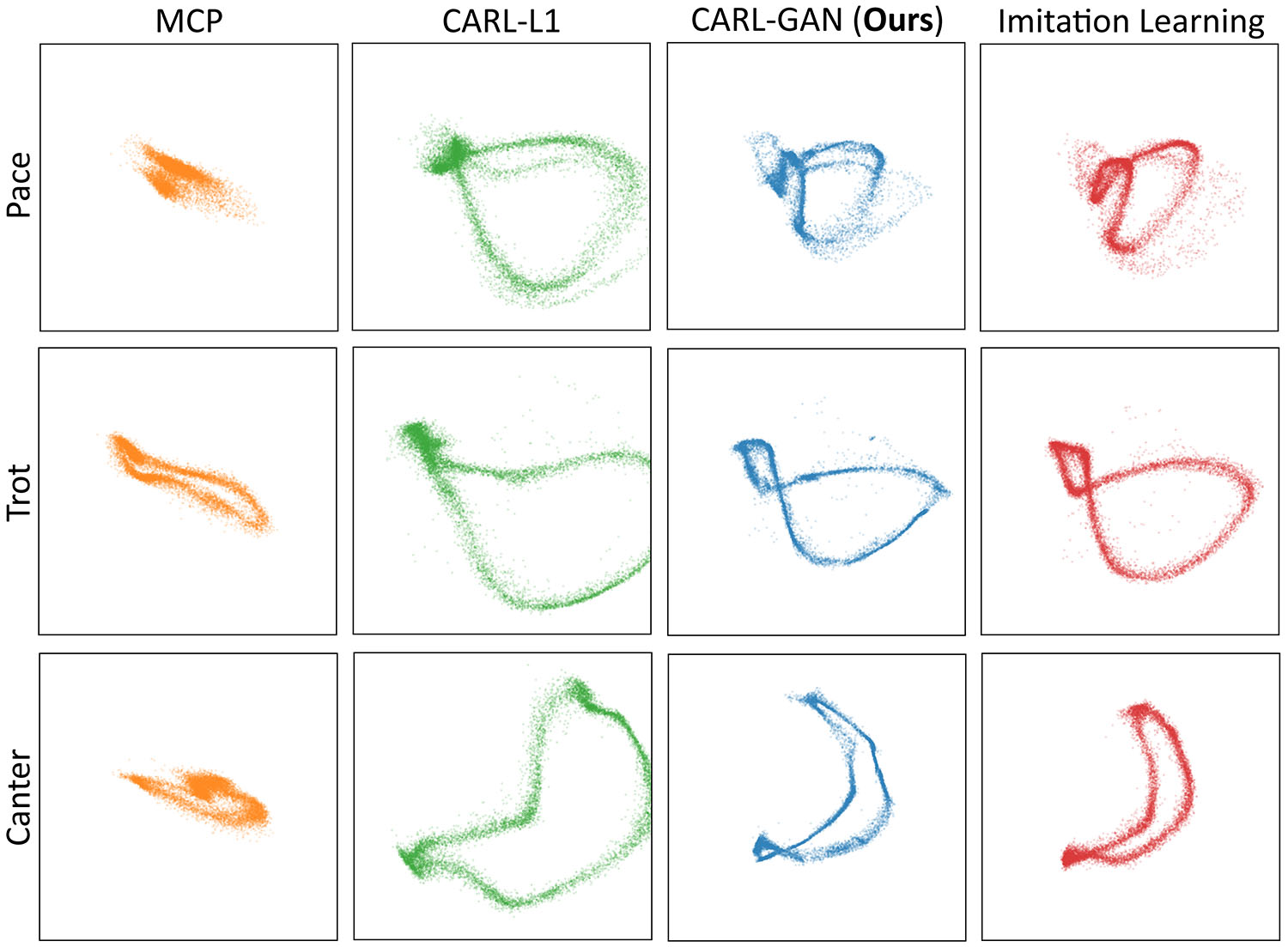}
  \vspace{-0.2cm}
  \caption{We apply a dimensional reduction technique (PCA) to visualize the action distribution in 2D. The visualization for each gait pattern forms a circular shape manifold corresponding to the agent's cyclic motion.}
  \label{fig:latent_space_exploration}
  \vspace{-0.35cm}
\end{figure}

\subsection{Action Space Exploration}
\label{sec:latent_space_exploration}
We find it useful to visualize the action distribution for assessing the quality of motions.
To this end, we collect millions of data points consisting of the agent's state in different gait patterns,
and apply standard principal component analysis (PCA) to visualize the data in 2D. The red dots in Figure \ref{fig:latent_space_exploration} represent the action distribution obtained from the imitation learning process. MCP (orange dots) produces a different distribution, which provides a visual clue on why it fails to learn the mapping between high-level user control and the extracted action distribution. CARL-L1 (green dots) preserves the shape of the action distribution, but the manifold appears to be expanded. Ours (blue dots) successfully captures the original action distribution, which indicates the effectiveness of the GAN Control Adapter in translating high-level user controls to the action distribution.
\raggedbottom

\begin{table}[]
    \caption{Intersection over Union (IoU) of an agent's end-effector while performing pace, trot, and canter.}
    \vspace{-0.25cm}
    \centering
    \begin{tabular}{ |c|c|c|c|c| } 
    \Xhline{2\arrayrulewidth}
    \textbf{Gait} & \textbf{Leg} & \textbf{MCP} & \textbf{CARL-L1} & \textbf{CARL-GAN} \\
    
    \Xhline{2\arrayrulewidth}
         & L.Front & 0.03 $\pm$ 0.01 & 0.04 $\pm$ 0.01 & \textbf{0.09  $\pm$ 0.01}\\
         & R.Front & 0.09 $\pm$ 0.02 & 0.14 $\pm$ 0.03 & \textbf{0.15  $\pm$ 0.02}\\
    Pace & L.Rear  & 0.35 $\pm$ 0.02 & 0.37 $\pm$ 0.02 & \textbf{0.42  $\pm$ 0.01}\\
         & R.Rear  & 0.02 $\pm$ 0.04 & \textbf{0.58 $\pm$ 0.02} & 0.51 $\pm$ 0.01\\
         & Avg.    & 0.12 $\pm$ 0.01 & 0.28 $\pm$ 0.02 & \textbf{0.29  $\pm$ 0.01}\\
    \Xhline{2\arrayrulewidth}
         & L.Front & 0.34 $\pm$ 0.02 & \textbf{0.65 $\pm$ 0.01} & 0.55 $\pm$ 0.01\\
         & R.Front & 0.34 $\pm$ 0.01 & 0.45 $\pm$ 0.04 & \textbf{0.61  $\pm$ 0.05}\\
    Trot & L.Rear  & 0.11 $\pm$ 0.02 & 0.17 $\pm$ 0.01 & \textbf{0.44  $\pm$ 0.03}\\
         & R.Rear  & 0.16 $\pm$ 0.01 & 0.49 $\pm$ 0.01 & \textbf{0.62  $\pm$ 0.01}\\
         & Avg.    & 0.24 $\pm$ 0.03 & 0.44 $\pm$ 0.02 & \textbf{0.56  $\pm$ 0.03}\\
    \Xhline{2\arrayrulewidth}
          & L.Front & 0.37 $\pm$ 0.02 & 0.41 $\pm$ 0.02 & \textbf{0.70 $\pm$ 0.03}\\
          & R.Front & 0.19 $\pm$ 0.01 & 0.29 $\pm$ 0.02 & \textbf{0.77 $\pm$ 0.03}\\
    Canter & L.Rear & 0.33 $\pm$ 0.03 & 0.47 $\pm$ 0.03 & \textbf{0.69 $\pm$ 0.04}\\
          & R.Rear  & 0.21 $\pm$ 0.01 & 0.28 $\pm$ 0.02 & \textbf{0.32 $\pm$ 0.03}\\
          & Avg.    & 0.27 $\pm$ 0.02 & 0.36 $\pm$ 0.02 & \textbf{0.62 $\pm$ 0.03}\\
    \Xhline{2\arrayrulewidth}
    \end{tabular}
    \label{tab:velocity_quant_iou}
    \vspace{-0.2cm}
\end{table}

\subsection{Kinematic Controller + DRL}
\label{sec:kcdrl}
\begin{figure}[b!]
  \centering
  \vspace{-0.4cm}
  \includegraphics[width=.95\linewidth]{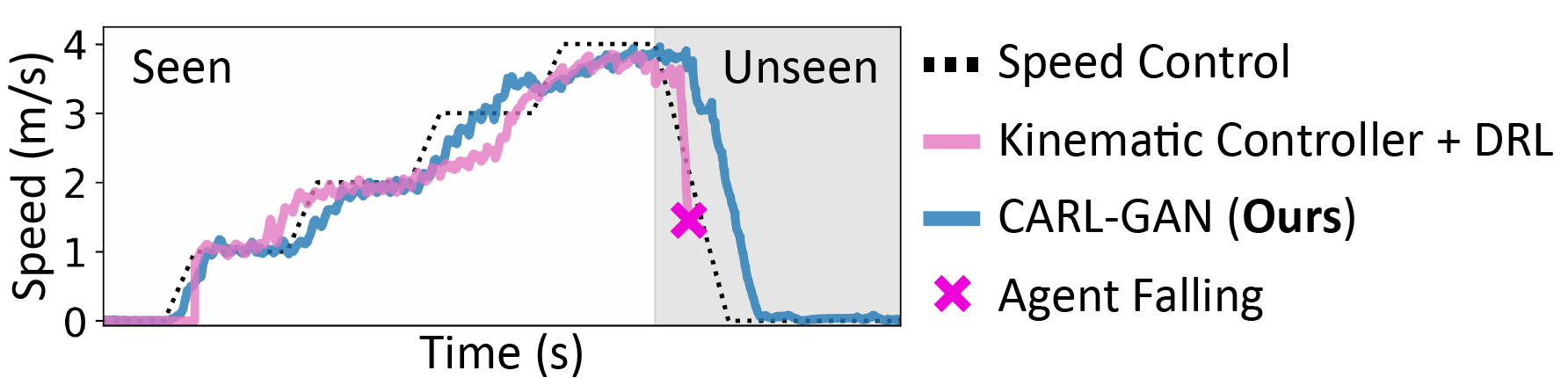}
  \vspace{-0.3cm}
  \caption{An example of speed control comparison given seen and unseen transitions.}
  \label{fig:kcdrl}
  \vspace{-0.7cm}
\end{figure}

Recent works from ~\cite{park2019learning, bergamin2019drecon} provide an end-to-end solution by manipulating the physics-based controller through a kinematic controller, i.e., Kinematic Controller + DRL. This approach can produce an accurate controller, high-quality motion, and is responsive, but requires access to action labels or a kinematic controller. In contrast, our method does not have such requirements, since we enable the controllable property by directly imposing high-level control to the physics-based controller. However, this causes an inherent problem of the distributional shift, which we address using GAN. 

Given a fixed control setting, the Kinematic Controller (MANN [Zhang et. al 2018]) + DRL is equivalent to our imitation learning, which produces high quality motion as highlighted by the manifold in the last column of Figure \ref{fig:latent_space_exploration}. As a comparison, we ask each controller to follow a set of speed control (dotted line in Figure \ref{fig:kcdrl}), from which the corresponding action labels are derived for the Kinematic Controller (e.g., 0 m/s for stand and 4 m/s for canter). When performing an unseen transition (4 m/s to 0 m/s), the Kinematic Controller + DRL fails to maintain balance and collapses to the ground due to the lack of such motion in the training data, as highlighted by the pink line in Figure ~\ref{fig:kcdrl}. To include all of the possible transitions, the training would require more reference motion data, which leads to exponentially longer training time, as suggested by ~\cite{park2019learning} in their ablation study. Despite using more reference data and training for a longer time, there is no guarantee that the controller would converge. In summary, via GAN, our method does not require access to either action labels or kinematic controllers and requires less reference motion data to model the gait transitions, therefore converges faster.
\begin{figure}[t!]
  \centering
  \includegraphics[width=.7\linewidth]{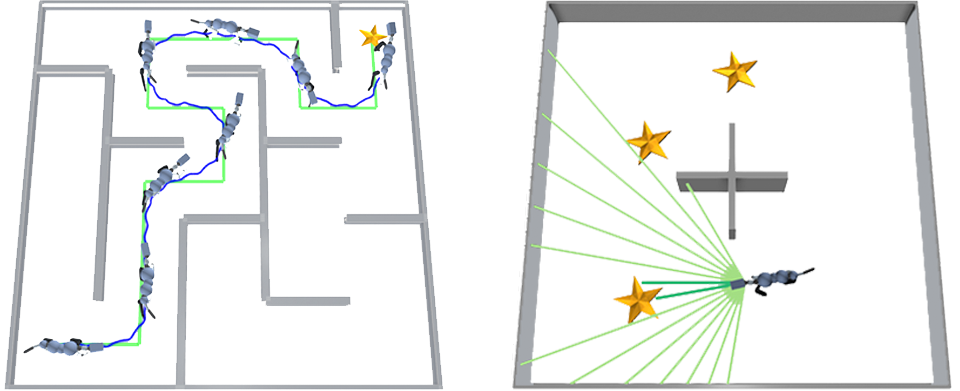}
  \vspace{-0.2cm}
  \caption{Attaching path-finding (left) and ray-sensor (right) navigation modules to the controller.} 
  \label{fig:application_navigation}
  \vspace{-0.25cm}
\end{figure}
\section{Applications}
\subsection{Modular Control}
Because the controller is trained to accurately follow the user's speed and heading controls, it becomes fairly straightforward to add complex navigation capability. The fact that the locomotion controller is decoupled from the navigation module, allows us to reuse pre-trained policy networks without compromising the desirable properties of the locomotion controller in complex navigation scenarios.
Consider a navigation task where the controller needs to collect valuable objects scattered within a maze. In Figure \ref{fig:application_navigation} (left), we implement a path-finding algorithm that reads the maze as a 2D map and outputs an optimal path for the controller to follow. In another task depicted in Figure \ref{fig:application_navigation} (right), the controller needs to rely on its sensors and observations to locate and collect the valuable objects. For this problem, we attach a ray-sensor and implement a classical navigation module. A simple translator module then converts the path into sequences of speed and heading controls. For instance, turning 90 degrees left at 1 m/s is represented as $c_{\text{high}} = (1,0.5\pi)$. The result is an autonomous agent that can efficiently solve the navigation task by following the sequence of controls.

\subsection{Reacting to External Perturbations}
\begin{figure}[t!]
  \centering
  \includegraphics[width=0.99\linewidth]{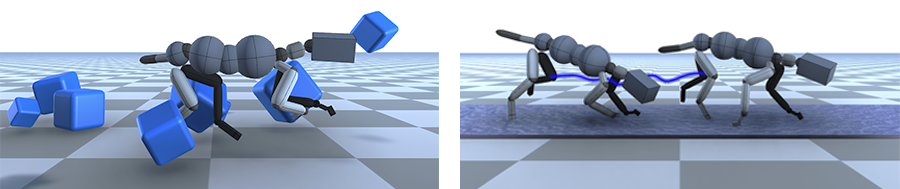}
  \vspace{-0.2cm}
  \caption{Our physics-based controller can produce meaningful reactions to randomly thrown boxes (left), and slippery terrain (right).} 
  \label{fig:application_perturbation}
  \vspace{-0.4cm}
\end{figure}

Our physics-based controller can effectively model the agent's interaction with a dynamically changing environment. To demonstrate this, we expose our agent to multiple unexpected external perturbations. The first perturbation involves boxes with various volumes and density from random directions. The second perturbation is a slippery terrain with the ground's friction reduced by 90\%. Lastly, we introduce a rotating and tilting terrain (in the video). Our experiment shows that the controller can react to the given perturbations, maintaining its balance while moving naturally and learn the importance of maintaining its height without being explicitly trained with the corresponding reward function (see Figure. \ref{fig:application_perturbation}).
\section{Conclusion}

In this paper, we present CARL, a quadruped agent that can respond naturally to high-level controls in dynamic physical environments. The three-stage process begins with a low-level imitation learning process to extract the natural movements perceived in the authored or captured animation clips. The GAN Control Adapter maps the high-level directive controls to action distributions that correspond to the animations. Further fine-tuning the controller with DRL enables it to recover from external perturbations while producing smooth and natural actions.
The GAN Control Adapter proves instrumental in enabling the controller to respond to high-level user controls, all while complying with the original action distribution, as demonstrated in the experiment where the GAN Control Adapter was removed or replaced with a simple L1 adapter. We show the usefulness of our method by attaching navigation modules over the controller to enable it to operate autonomously for tasks such as traversing through mazes with goals. In addition to that, we also create an animation clip, where the agent-environment interaction changes dynamically. Equipped with natural movements, controllable, adaptive properties, our approach is a powerful tool in accomplishing motion synthesis tasks that involve dynamic environments.

In the future, we would like to investigate on imposing constraints over the manifold's morphology, such as shape and size, for controlling the movement style and transitions. Another interesting direction is to use a dynamic lambda to regularize the fine-tuning stage, as a fixed lambda may lead to a trade-off between realism and adaptiveness. We hope this research in combining the merits of GAN and DRL can inspire future research directions towards high realism controllers.

\begin{acks}
We wish to thank the anonymous reviewers for the insightful comments, Yi-Chun Chen for the help in producing the supplementary video, and our colleagues for the meaningful discussions.
\end{acks}

\bibliographystyle{ACM-Reference-Format}
\bibliography{main}
\end{document}